\documentclass[letterpaper, 10 pt, conference]{ieeeconf}  

\IEEEoverridecommandlockouts                              
\usepackage{cite}
\usepackage{float}
\usepackage[pdftex]{graphicx}
\usepackage{amsmath,amssymb}
\usepackage{array}
\usepackage{url,hyperref}

\usepackage{dblfloatfix}
\usepackage{multirow}
\usepackage{booktabs}
\usepackage{subcaption}
\usepackage{xcolor}
\usepackage{etoolbox}

\makeatletter
\patchcmd{\@makecaption}
{\\}
{.\ }
{}
{}
\makeatother

\centerfigcaptionsfalse

	\title{	Real Time Lidar and Radar High-Level Fusion for Obstacle Detection and Tracking with evaluation on a ground truth}

\author{Hatem Hajri and Mohamed-Cherif Rahal $^{*}$
	\thanks{}
	\thanks{$^{*}$Institut VEDECOM, 77, rue des Chantiers – 78000 Versailles, France.
		{Email: \tt\small firstname.name@vedecom.fr}}%
	\thanks{}}%

\begin{document}
	
\maketitle
\thispagestyle{empty}
\pagestyle{empty}

\begin{abstract}
Both Lidars and Radars are sensors for obstacle detection. While Lidars are very accurate on obstacles positions and less accurate on their velocities, Radars are more precise on obstacles velocities and less precise on their positions. Sensor fusion between Lidar and Radar aims at improving obstacle detection using advantages of the two sensors. 
The present paper proposes a real-time Lidar/Radar data fusion algorithm for obstacle detection and tracking based on the global nearest neighbour standard filter (GNN). This algorithm is implemented and embedded in an automative vehicle as a component generated by a real-time multisensor software. The benefits of data fusion comparing with the use of a single sensor are illustrated through several tracking scenarios (on a highway and on a bend) and using real-time kinematic sensors mounted on the ego and tracked vehicles as a ground truth. 
\end{abstract}

\IEEEpeerreviewmaketitle
\section{Introduction}
Data fusion \cite{Hall04a,Mitchell:2007:MDF:1543556,Barshalom88a} has multiple benefits in the field of autonomous driving. In fact autonomous vehicles are often equipped with different sensors through which they communicate with the external world. A multisensor fusion takes advantages of each sensor and provides more robust and time-continuous informations than sensors used separately.

Several earlier works showed advantages of combining sensors such as Lidars, Radars and Cameras. For example, \cite{Blanctrackto} presents a Lidar-Radar fusion algorithm based on Kalman filter and shows how fusion improves interpretation of road situations and reduces false alarms. Subsequently \cite{bbbbb} uses Cramer-Rao lower bound to estimate performance of data fusion algorithms. The paper \cite{ttt} considers Lidar-Radar fusion with applications to following cars on highways. In order to test the performance of their fusion algorithm, authors of \cite{ttt} provide a study of mean square errors of relative distances and velocities in a highway tracking scenario using least squares polynomial approximation of sensors data as a ground truth. Another Lidar, Radar and Camera fusion approach based on evidence theory apppears in \cite{Chavez-Garcia2015} with applications to the classification and tracking of moving objects. More recently, \cite{new2018} focuses on fusion of multiple cameras and Lidars and presents tests on real world highway data to verify the effectiveness of the proposed approach.

The present paper is concerned with data fusion between Lidar and Radar. In vue of the state of the art, we can distinguish two different general fusion methods which were applied for Lidar and Radar: Kalman filter and evidence theory. We believe that approaches which apply one of these methods agree on the main steps. On the other hand, despite the previously mentionned works, the litterature still clearly lacks a quantitative comparison between these sensors outputs such as relative coordinates, velocities, accelerations etc and their fusion result in the presence of a ground truth at least on one obstacle. Because of the lack of a ground truth, authors of these papers were led to work with simulated data or manually manage real data in order to create a ground truth and evaluate results. \\ 
The present paper proposes a real-time high-level fusion algorithm between Lidar and Radar based on the GNN filter which in turn is based on Kalman filter. This algorithm is presented with several mathematical and implementation details which go along with it. The performance of this algorithm is evaluated while focusing on the main outputs of Lidar and Radar which are relative coordinates and velocities of obstacles. 
Evaluation is done using a ground truth methodology introduced in \cite{hhh}. For this, two synchronised autonomous cars which are prototypes of the autonomous vehicle of VEDECOM equipped with real-time kinematic (RTK) sensors will be used in tracking scenarios.  Figure 1 displays the sensor architecture of this prototype. It has five Lidars ibeo LUX with horizontal fields of view of 110$^o$ mounted such that they provide a complete view around the car. These Lidars send measurements to the central computation unit (Fusion Box) which performs fusion of measured features, object detections and tracks at the frequency of 25 Hz. A long-range Radar ARS 308 of frequency 15 Hz is mounted at the front of the car with a horizontal view of $-28^o,+28^o$. The vehicle is moreover equipped with a RTK sensor with precisions $0.02$ m on position and $0.02$ m/s on velocity and a CAN bus which delivers odometry informations.
\begin{figure}[H]
	\centering
	\includegraphics[width=6cm,height=3.5cm]{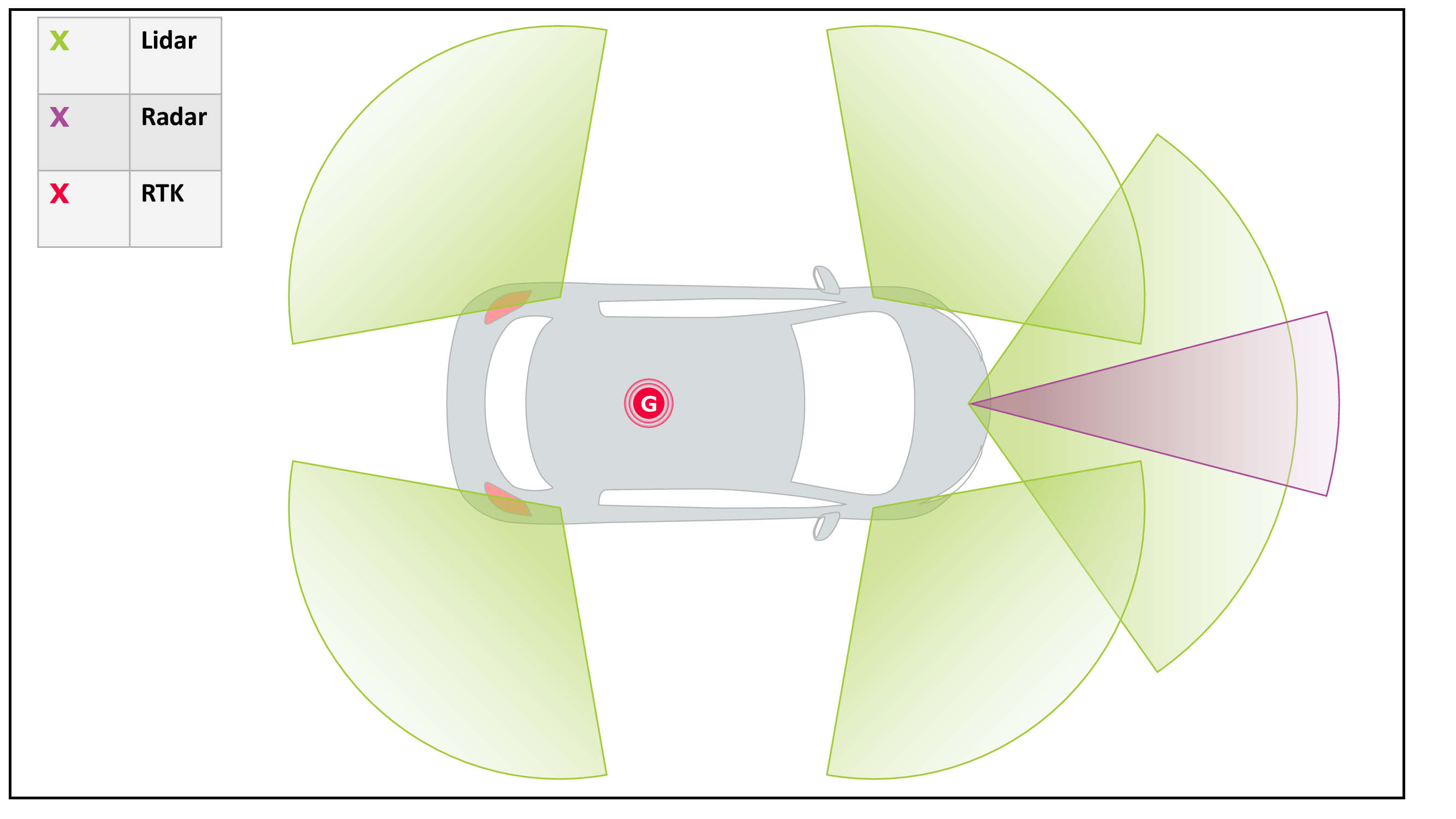}
	\caption{Sensor configuration: Five lidars, one radar and one RTK sensor}

\end{figure} 
The fusion algorithm uses informations from Lidar/Radar and the CAN bus. RTK sensors will be used for performance evaluation. 

\textbf{Organisation of the paper.} The content of this paper is as follows. Section \ref{Kalman} develops the fusion method used in the paper. Section \ref{implem} gives more details on the implementation and integration into vehicle of the proposed algorithm. Section \ref{expe} reviews the ground truth generation method introduced in \cite{hhh} and presents the experiments (car followings on highways/bends) carried out using the two vehicles to collect data. The ground truth is used to evaluate the mean square erros of Radars/Lidars and fusion measurements of relative positions and velocities of the obstacle vehicle. In addition several plots of temporal evolutions of these measurements are given showing interesting informations about their smoothness and unavailability periods. Results show advantages of data fusion comparing with one sensor.

\section{Kalman Filter for data fusion}\label{Kalman}
Consider a target (or state) which moves linearly in discrete time according to the dynamic $x(k)=F(k) x(k-1) +  v(k)$ and an observation $z$ of it by a sensor $S$ which takes the form $z(k)=H(k) x(k) +  w(k)$. Assume $v(k),k=1,\cdots$ and $w(k),k=1,\cdots$ are two independent centered Gaussian noises with covariances $P_S$ and $N_S$ respectively. Kalman filter is known to give an estimation of $x(k)$ when only $z(1),\cdots,z(k)$ are observed. From a mathematical point of view, the problem amounts to calculating the conditional mean $\mathbb E[x(k)|Z^k]$ where $Z^k=(z(1),\cdots,z(k))$. Introduce the notations $\hat x(i|j)=\mathbb E[x(i)|Z^j]$ which is the conditional mean of $x(i)$ knowing $Z^j$ and its conditional covariance $$P(i|j)=\mathbb E[(x(i) - \hat x(i|j)) (x(i) - \hat x(i|j))^T|Z^j].$$
Kalman filter has an explicit solution which is determined recursively.
Assume $x(0)$ is a Gaussian distribution with mean $\hat{x}(0|0)$ and covariance $P(0|0)$. Knowing the estimation $\hat{x}(k-1|k-1)$, $\hat{x}(k|k)$ is calculated following these two steps.

\noindent\textsc{(a) Prediction step.} Compute $\hat{x}(k|k-1)$ and $P(k|k-1)$ by:
\begin{eqnarray}
\hat{x}(k|k-1)&=&F(k)\hat{x}(k-1|k-1)\nonumber\\
P(k|k-1)&=&F(k)P(k-1|k-1)F^T(k) + P_S\nonumber\
\end{eqnarray}
This step requires the knowledge of $z(1),\cdots,z(k-1)$.

\noindent\textsc{(b) Update step.} When $z(k)$ becomes available the final solution is obtained as follows
\begin{eqnarray}
\hat{x}(k|k)&=&\hat{x}(k|k-1) + W(k) (z(k) - H(k)\hat{x}(k|k-1))\nonumber\\
P(k|k) &=& P(k|k-1) - W(k)S(k)W^T(k)\nonumber\
\end{eqnarray}
with 
\begin{eqnarray}
W(k)&=&P(k|k-1) H^T(k) S^{-1}(k)\ \ \text{(Kalman gain)}\nonumber\\
S(k)&=&N_S + H(k) P(k|k-1) H^T(k)\nonumber\
\end{eqnarray}
The matrix $S(k)$ is the covariance of the innovation $\nu(k)=z(k) -  H(k)\hat{x}(k|k-1)=z(k) - \hat{z}(k|k-1)$. The innovation measures the deviation between the estimates provided by the filter and the true observations. Its practical interest lies in the fact that (under the Gaussian assumption) the normalised innovation $q(k)=\nu^T(k) S(k)^{-1}\nu(k)$ is a $\chi^2$ distribution with $\text{dim}(z(k))$ degrees of freedom (see \cite{Barshalom88a,whyy}). As a consequence, with a high probability $\alpha$, the observation $z(k)$ associated with the object $x(k)$ belongs to the area
$$\{z: d^2=(z - \hat{z}(k|k-1))^T S(k)^{-1} (z - \hat{z}(k|k-1))\le \gamma\}$$
where $\alpha$ is such that $\mathbb P (\chi ^ 2 <\gamma) = \alpha$. This area is known as the validation gate. When several measurements $z$ are available for one specific object, the $\chi^2$ distribution makes it possible to identify those measurements which may correspond to the underlying object by applying a hypothesis test as follows: If there are observations in the validation gate choose $z$ which minimizes $d^2$ as the only valid observation. Otherwise no observation is associated. This solution is known as the nearest neighbour filter. The solution which consists in averaging all obsevations which fall in the validation gate is known as the probabilistic data association filter (see \cite{Barshalom88a} for more background on probabilistic filters). In practice, which is the case for Radar/Lidar, several tracked objects and new measurements can be available simultaneously. The task of associating new measurements with the underlying observed objects is known as the data association problem. This problem is dealt with in Section \ref{implem} by first solving a global optimization problem and second updating the list of objects using the nearest neighbour filter. The obtained filter is known as the GNN filter.

In the present paper each obstacle (old state or new observation) is represented as a vector of its relative coordinates and velocities $[x\ y\ v_x\ v_y]^T$. It is assumed that obstacles move according to the constant velocity motion model: if $\Delta$ denotes the elapsed time between the $k$ and $k+1$ sensor data emissions, the matrices $F$ and $H$ describing the state and observation evolutions are given by
$$F(k)=
 \begin{pmatrix} 
 	1 & 0 & \Delta & 0\\
 	0 & 1& 0&\Delta\\
 	0 &0& 1& 0\\
 	0 &0 &0 &1\\ 
 \end{pmatrix},\ \ H(k)=I_4$$

The content of this paper can be adapted to other motion models such as the constant acceleration motion model. We choose the constant velocity model for two reasons. First Lidars do not send informations about accelerations of obstacles. Second the constant velocity model requires less unkown parameters to estimate than the constant acceleration model. 
\section{Real time implementation}\label{implem}
This section gives details of the implementation of the real-time fusion module between the two sensors Lidar ibeo LUX and Radar ARS 308. This module was first implemented in Matlab/Simulink and then embedded in the automative vehicle as a component generated by the software RTMaps (Real Time, Multisensor applications) \cite{unknown}. The RTMaps software is widely used for real-time applications in mobile robotics as it allows to synchronise, record and replay data from different sensors. The proposed fusion module is an iterative algorithm which is run as soon as a sensor data arrives. Simulink delay functions offer a tool for this kind of problems as they allow to store the last value of a module. The following Fig. \ref{figsimA} shows the Simulink diagram used to generate the fusion module. 
\begin{figure}[H]
	\centering
	\includegraphics[width=8cm]{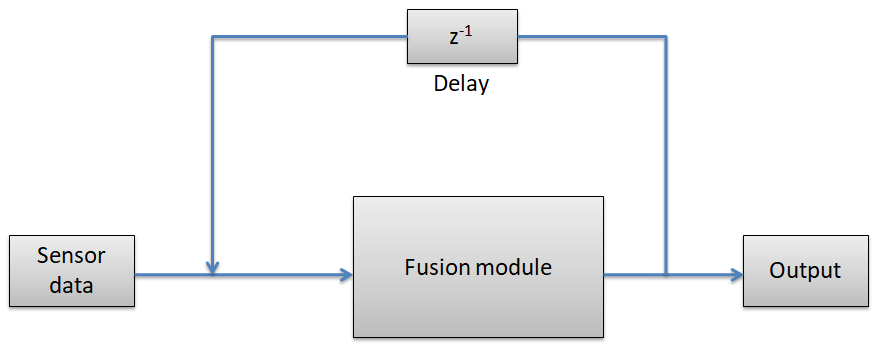}
	\caption{The Simulink diagram}
	\label{figsimA}
\end{figure}
After a data by a sensor $S_k$ (Lidar or Radar) is received at time $t_k$, this algorithm outputs a fused set of obstacles $FO_k$. The main characteristics of each obstacle (other outputs such as identity, age and so on will not be detailled) are 1) a vector $[x,y,v_x,v_y]^T$ where $x,y$ (resp. $v_x,v_y$) are the relative (with respect to the ego-vehicle frame) coordinates (resp. velocities) of the object and 2) an uncertainty $4 \times 4$ matrix of the object. The input of the algorithm at time $t_k$ is the previous fused set of obstacles $FO_{k-1}$, the linear and angular velocities of the ego-vehicle at time $t_k$ and the sensor data consisting in relative coordinates $x, y$ and speeds $v_x, v_y$ of the detected obstacles all at time $t_k$. The angular and linear velocities are available on the CAN bus. Since this sensor does not generate data at the same moments as Lidar or Radar, in practice an approximation value of these quantities at time $t_k$ are taken (for example the last known ones before $t_k$). 

The fused list is initialized when the first sensor data arrives. For each object in this list, the vector $[x,y,v_x,v_y]^T$ of relative coordinates and velocities given by the sensor is stored. The uncertainty of each object is initialized to the corresponding sensor's uncertainty $N_S$. 

Assume known the last fused list of objects $FO_{k-1}$ at time $t_{k-1}$ updated after reception of data by the sensor $S_{k-1}$ and assume a new sensor data arrives at time $t_k$ by the sensor $S_k$. The update $FO_{k}$ of $FO_{k-1}$ follows these steps.\\      
\noindent\textbf{(a) Tracking.} Call $\mathcal R_{k-1}$ and $\mathcal R_k$ the ego-vehicle frames at times $t_{k-1}$ and $t_k$. Each object $o^{k-1}=[x^{k-1},y^{k-1},v_x^{k-1},v_y^{k-1}]^T\in FO_{k-1}$ with uncertainty $I^{k-1}$ is first tracked in the frame $\mathcal R_{k-1}$ under the constant velocity hypothesis. Then it is mapped to the new frame $\mathcal R_{k}$ by a rotation of its relative coordinates and relative velocities. Call $\omega$ and $v$ the instantaneous angular and linear velocities of the ego-vehicle at time $t_k$ and define the following estimates of the cap angle and travelled distance $\theta=\omega \times \Delta, d=v\times \Delta$ where $\Delta=t_k-t_{k-1}$. Call $o^{k-1,t}=[x^{k-1,t},y^{k-1,t},v_x^{k-1,t},v_y^{k-1,t}]^T$ the new object and $I^{k-1,t}$ its uncertainty in $\mathcal R_k$. More explicitly, the following identities hold: $[x^{k-1,t},y^{k-1,t},v_x^{k-1,t},v_y^{k-1,t}]^T$
$$
=
\begin{pmatrix} 
R_{\theta} & 0 \\
0 & R_{\theta} 
\end{pmatrix} \begin{pmatrix} 
x^{k-1}+\Delta v_x^{k-1} - d \cos(\theta) \\
y^{k-1}+\Delta v_y^{k-1} - d \sin(\theta)  \\
 v_x^{k-1}\\
 v_y^{k-1}
\end{pmatrix}$$

and

$$I^{k-1,t}=\begin{pmatrix} 
R_{\theta} & 0 \\
0 & R_{\theta} 
\end{pmatrix} (F I^{k-1} F^T+P_{S_k}) \begin{pmatrix} 
R_{\theta} & 0 \\
0 & R_{\theta} 
\end{pmatrix}^T$$

\noindent where $R_{\theta}$ is the rotation matrix of angle $\theta$ and $F$ is given in the previous section.\\
\noindent\textbf{(b) Association.} In this setp, each new sensor measurement is associated with at most one tracked object that corresponds best to it. For this, the classical Mahalanobis distance will be used as a similarity measure. This distance is defined for any tracked object $o^{k-1,t}$ with uncertainty $I^{k-1,t}$ in $\mathcal R^k$ and new observation $z^i$ by $$d^2_{k,i}=(o^{k-1,t}-z^i)^T(I^{k-1,t})^{-1}(o^{k-1,t}-z^i).$$
The association problem can be reformulated in the following optimization form: find the set $(c_{k,i})$ which minimizes $\sum_{k,i} c_{k,i} d^2_{k,i}$ subject to the constraints $\sum_{k} c_{k,i}=1$ for each $i$, $\sum_{i} c_{k,i}=1$ for each $k$ and $c_{k,i}\in\{0,1\}$ for all $k$ and $i$. A well known solution to this problem is given by the Hungarian algorithm \cite{Munkres1957}. This algorithm was implemented in Matlab/Simulink and subsequently used in the fusion module. Notice if there are less tracked objects than new observations, a tracked object is associated with exactly one observation and if not it may or may not be associated.\\            
\noindent\textbf{(c) List update.} The fused list of objects is updated as follows. First for a tracked object $o^{k-1,t}$ with uncertainty $I^{k-1,t}$ which is associated with an observation $z^i$, the latter is accepted as an observation of $o^{k-1,t}$ if it falls in the validation gate arround $o^{k-1,t}$ that is if $d^2_{k,i}<\gamma$ with $\gamma$ a quantile of order $0.9$ of the $\chi^2$ distribution with $4$ degrees of freedom. In this case both $o^{k-1,t}$ and $I^{k-1,t}$ are updated as in Kalman correction step based on the observation $z^i$. If the object is not associated with an observation, it is removed from the list. Another possibility is to continue tracking absent obstacles for a while. However we did not choose this option since Lidars and Radars already track absent obstacles. Finally new observations that are not associated with tracked objects are added to the list as for the first fused list. To summarize, the new list $FO_k$ is composed of tracked objects associated with new observations and updated by Kalman filter and new observations not associated with tracked objects. 
\section{Experiments and results}\label{expe}
The last section is devoted to the generation of ground truth for the evaluation of the proposed data fusion algorithm (denoted Fusion in brief) and for comparisons between Lidar, Radar and Fusion. The focus will be on relative positions/velocities $x, y,vx, vy$ of the target vehicle.\\
\textbf{Ground truth generation \cite{hhh}.} To generate the ground truth, we used two synchronised autonomous vehicles. The ego vehicle is equipped with Lidar and Radar and both vehicles are equipped with RTK sensors. The idea behind this generation process is that given two moving points $A$ and $B$ represented by their positions and velocities in a global frame $R$ and knowing the angular velocity of $A$, it is possible to deduce the positions and velocities of $B$ in the moving frame of $A$ (composition of movements formula). In practice, these informations are used: 

\begin{itemize}
	\item[(1)] global coordinates and velocities in the reference frame $\mathcal R_0$ of RTK sensors of the two vehicles during experiments all obtained from the RTK sensors. 
	\item[(2)] heading of the ego vehicle in $\mathcal R_0$ during experiments obtained from the RTK sensor mounted on this vehicle.

\end{itemize} 
These informations in combination with the composition of movements formula provide ultra-precise estimations of the relative positions/velocities of the target vehicle in the ego vehicle frame. In fact, at each time $t$, informations (1) give the relative coordinates and speeds with the ego-vehicle identified to a point (that is without consideration of its heading). A rotation of angle given by the heading at time $t$ (obtained from (2)) gives the desired estimations. In order to perform comparison, one has to find the target car characteristics $(x_t,y_t,vx_t,vy_t)$ sent by Lidar/Radar/Fusion at a given time $t$. For this, the nearest obstacle to the ground truth position at time $t$ which is non static was considered as the target car viewed by Lidar/Radar/Fusion. This method gives the desired obstacle most all the time. Since sensors have different frequencies, linear interpolation was used to get an estimate of any quantity (position/velocity) which is not available at a given time. \\
\textbf{Evaluation of the fusion algorithm.} In order to estimate the uncertainty matrices $P_S$ and $N_S$ for both Lidar and Radar involved in the algorithm, we used a ground truth collected during 10 minutes. We set all off-diagonal entries of these matrices to $0$ and tolerate more error than the obtained values. To evaluate the fusion algorithm, we generated new data in two challenging circuits: car followings on a highway and a highly curved bend. In each case, the same car-following scenario is repeated seven times. For each case, $4$ figures displaying variations of the relative coordinates and velocities $x,y,vx,vy$ in one scenario are shown along with the ground truth. We choose one scenario among seven because of space limitations. The two plots at the bottom (green and black refering to Radar and RTK) are the true ones (without translation). For better visualization, those at the midlle (red and black refering to Lidar and RTK) and the top (blue and black refering to Fusion and RTK) correspond to the true ones + offset and true ones + 2*offset. The offset is specified with each figure's title. Radar/Lidar/Fusion points are represented by small squares joined by lines with the same colors and RTK points are represented by continuous curves (linear interpolation of the values). In addition, two tables representing the mean square errors (MSE) of $x,y,vx,vy$ for Radar/Lidar/Fusion are given for all the seven car following experiencies. These errors are calculated with respect to the RTK output considered as a ground truth according to the formula
$$\text{MSE on q} = \frac{1}{N}\sum_{t=1}^N (\text{qSensor}_t - \text{qRTK}_t)^2$$
In this formula, the notation qSensor with $q\in\{x,y,vx,vy\}$ and sensor $\in\{\text{Radar,Lidar,RTK, Fusion}\}$ refers to $q$ of the target as seen by sensor, qSensor$_t$ is the value of qSensor at time $t$ and $N$ is the number of samples.\\
\textbf{Runtime performance.} Multiple tests show that the fusion algorithm is able to treat 50 obstacles in less than 15 microsecondes. This period is negligible compared to periods of Lidar and Radar making the algorithm suitable for fusion in real time. 
\subsection{Car-following on a highway} 
The first scenario is a tracking, shown in Figure 3, of the target vehicle on a highway at high speed (between $90$ and $100$ km/h). We repeated the same experience on the same portion of the highway seven times and obtained a record of more than 6 minutes.  
\begin{figure}[H]
	\centering
	\includegraphics[width=7.5cm,height=4.5cm]{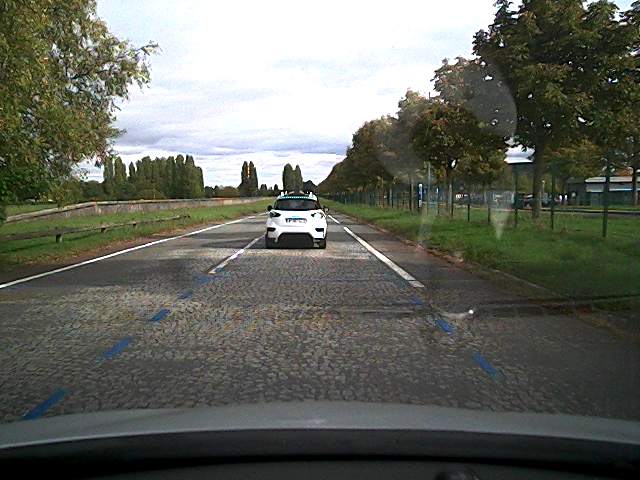}
	\caption{Tracking on a highway.}
\end{figure}
The next figures display respectively the variations of $x,y$ (in metre) $vx,vy$ (in metre/second) as a function of time (in second) of the target vehicle by Radar/Lidar/Fusion and RTK in one experience among the seven.
\begin{figure}[H]
	\centering
	\includegraphics[width=8cm,height=5.3cm]{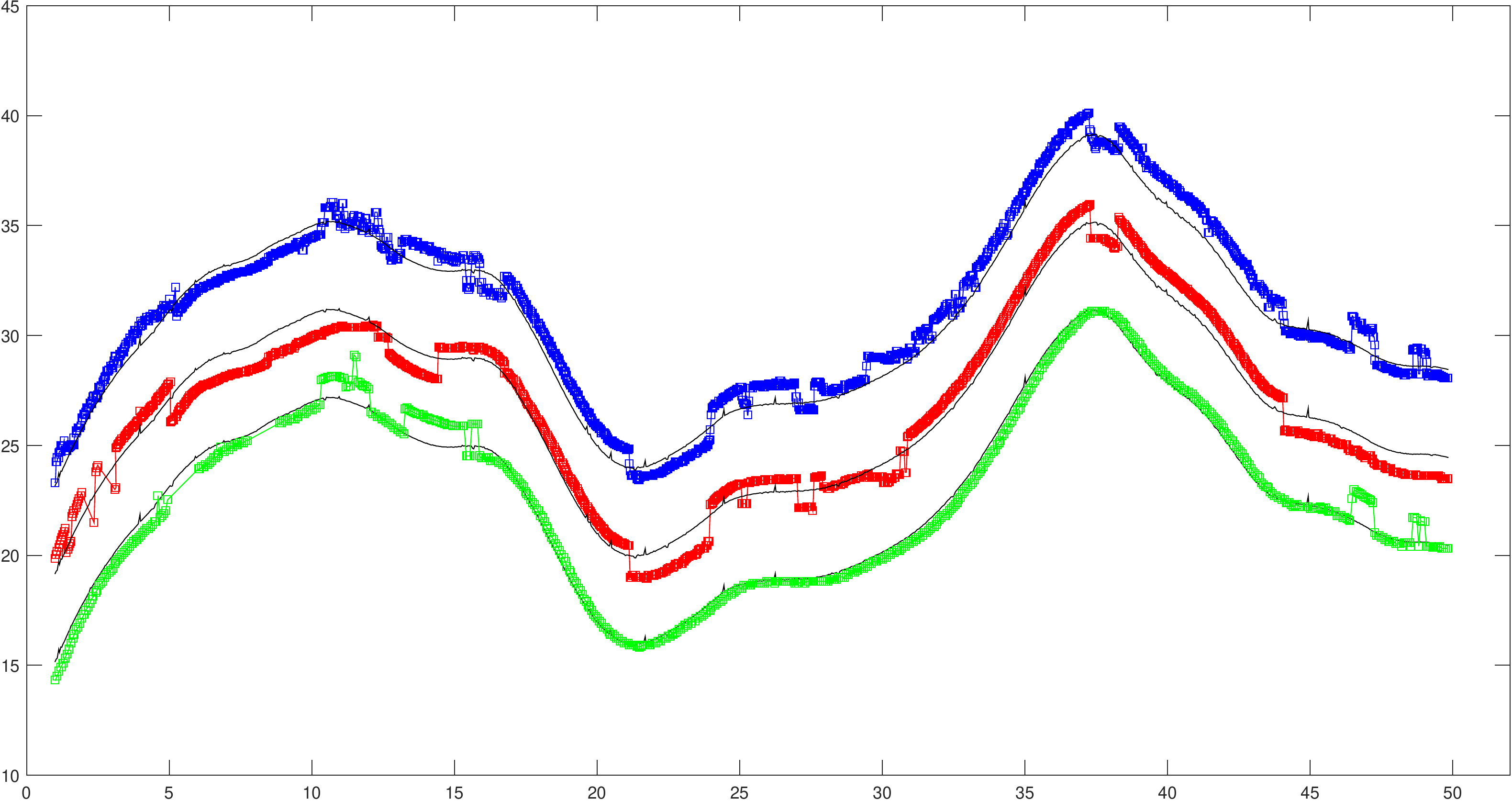}
	\caption{variations of $x$ (in m) with Radar (green)/Lidar (red)/Fusion (blue) in comparison with RTK (black) as a function of time (in s). Offset=4.}

\end{figure}
\begin{figure}[H]
	\centering
	\includegraphics[width=8cm,height=5.3cm]{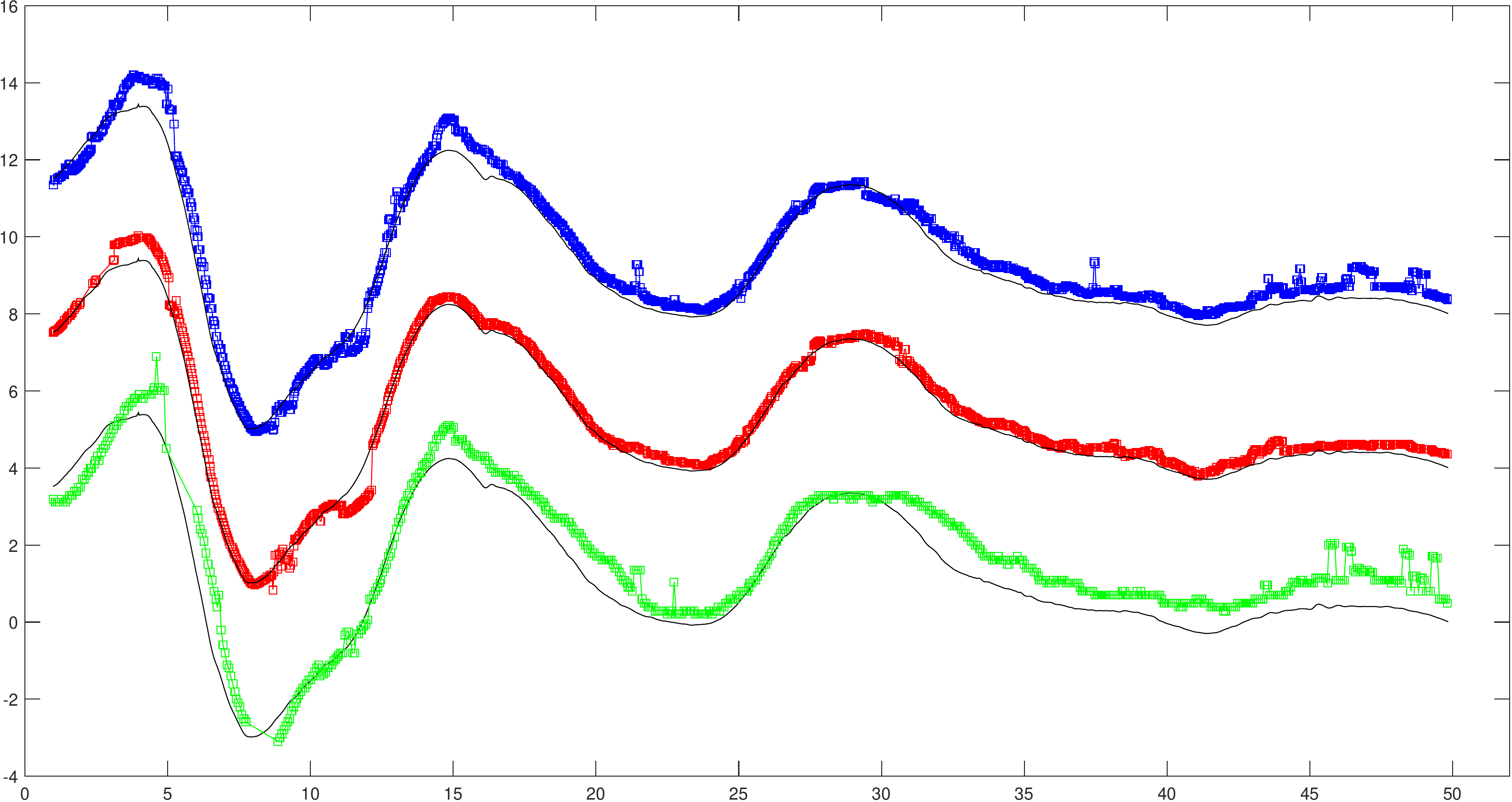}
	\caption{variations of $y$ with Radar/Lidar/Fusion. Offset=4.}

\end{figure}
\begin{figure}[H]
	\centering
	\includegraphics[width=8cm,height=5.3cm]{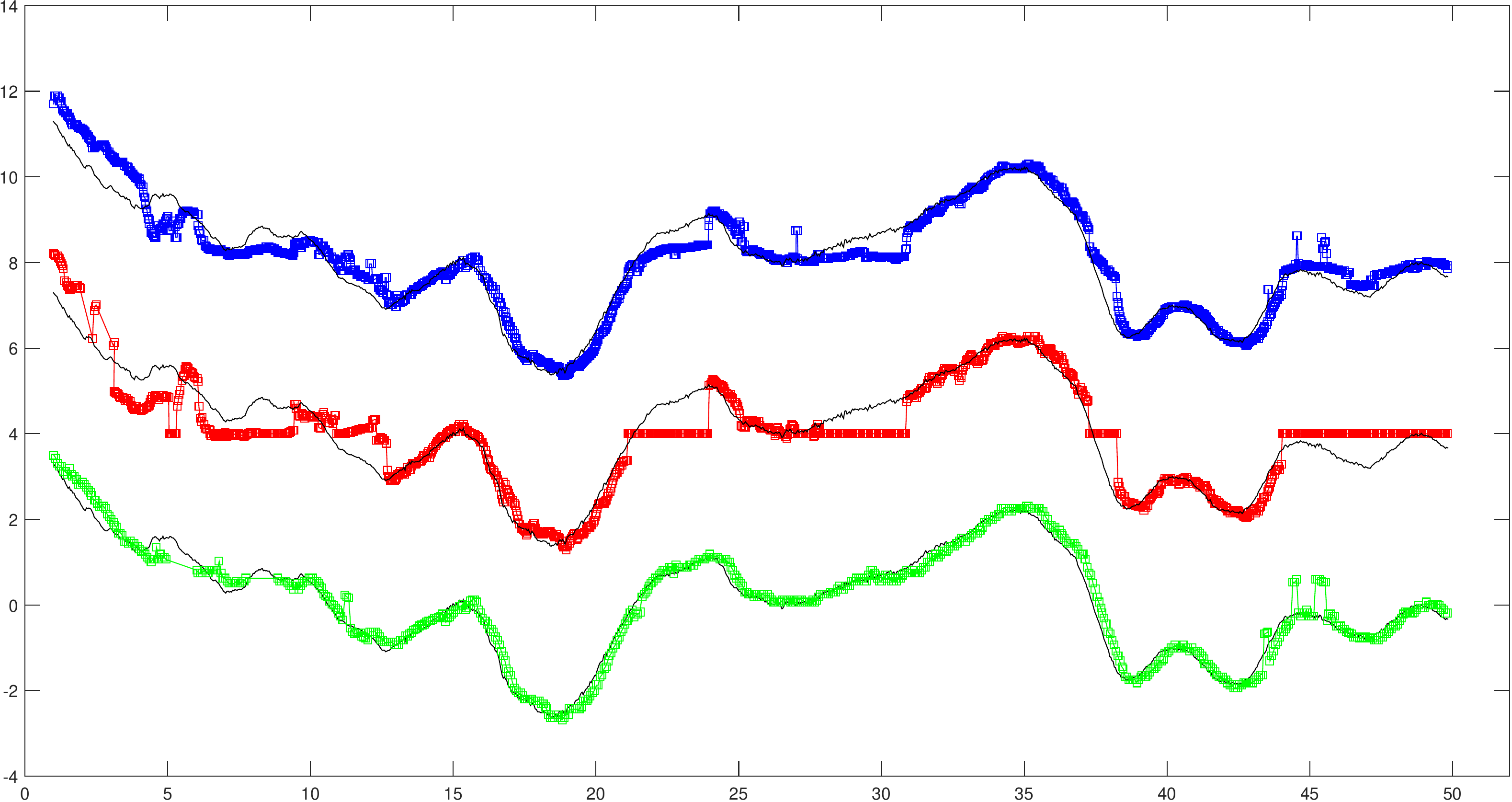}
	\caption{variations of $vx$ with Radar/Lidar/Fusion. Offset=4.}

\end{figure}
\begin{figure}[H]
	\centering
	\includegraphics[width=8cm,height=5.3cm]{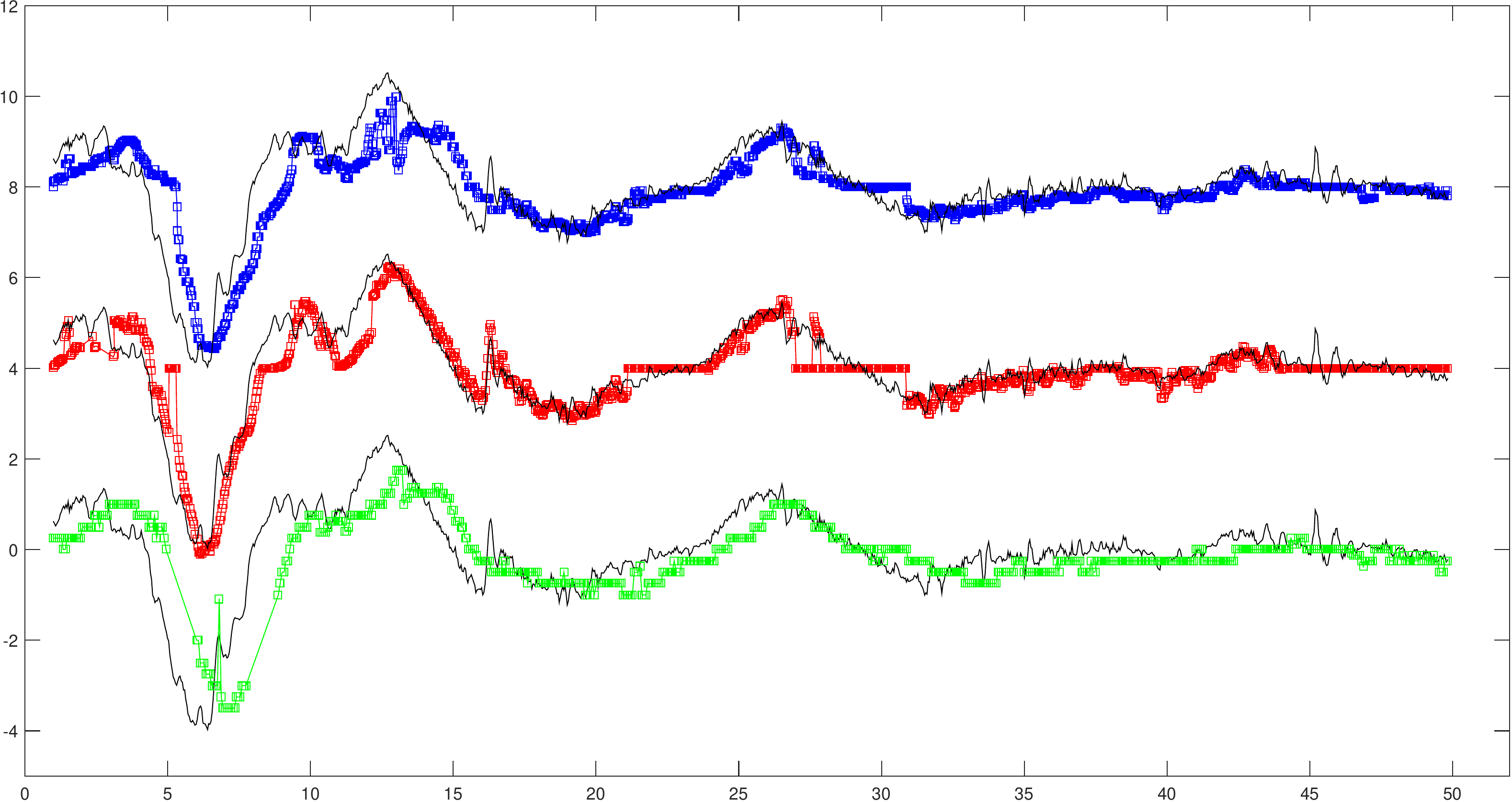}
	\caption{variations of $vy$ with Radar/Lidar/Fusion. Offset=4.}

\end{figure}
Table \ref{tab3} shows the mean square erros (MSE) on $x,y,vx,vy$ by Radar,Lidar,Fusion corresponding to the seven car-following experiences. 
\begin{table}[H]
	\centering
	\begin{tabular}{|l||l|l|l|}
		\hline  & Radar &  Lidar &  Fusion  \\
		\hline  MSE on $x$  & $0.33$ & $0.55$  &$0.39$\\
		\hline  MSE on $y$ & $0.43$ & $0.16$ & $0.21$  \\
		\hline  MSE on $vx$ & $0.15$ & $0.28$ & $0.19$  \\
		\hline  MSE on $vy$ & $0.25$ & $0.31$ & $0.27$  \\
		\hline
	\end{tabular} 
	\caption{Table of MSEs of $x,y,vx,vy$ by /Radar/Lidar/Fusion.}
	\label{tab3}
\end{table}
\subsection{Car-following on a bend} 
The second scenario is a tracking, shown in Figure 8, on a succession of highly cruved bends. We repeated the same experience seven times getting a record of more than 6 minutes.  
\begin{figure}[H]
	\centering
	\includegraphics[width=7.5cm,height=4.1cm]{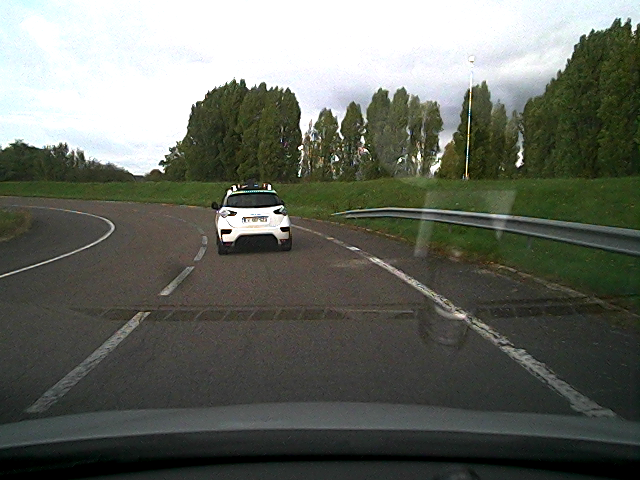}
	\caption{Tracking on a bend.}
\end{figure} 
The following Figure 9 shows the trajectory in the world frame of the ego vehicle obtained from the RTK sensor.  
\begin{figure}[H]
	\centering
	\includegraphics[width=8.5cm,height=4.2cm]{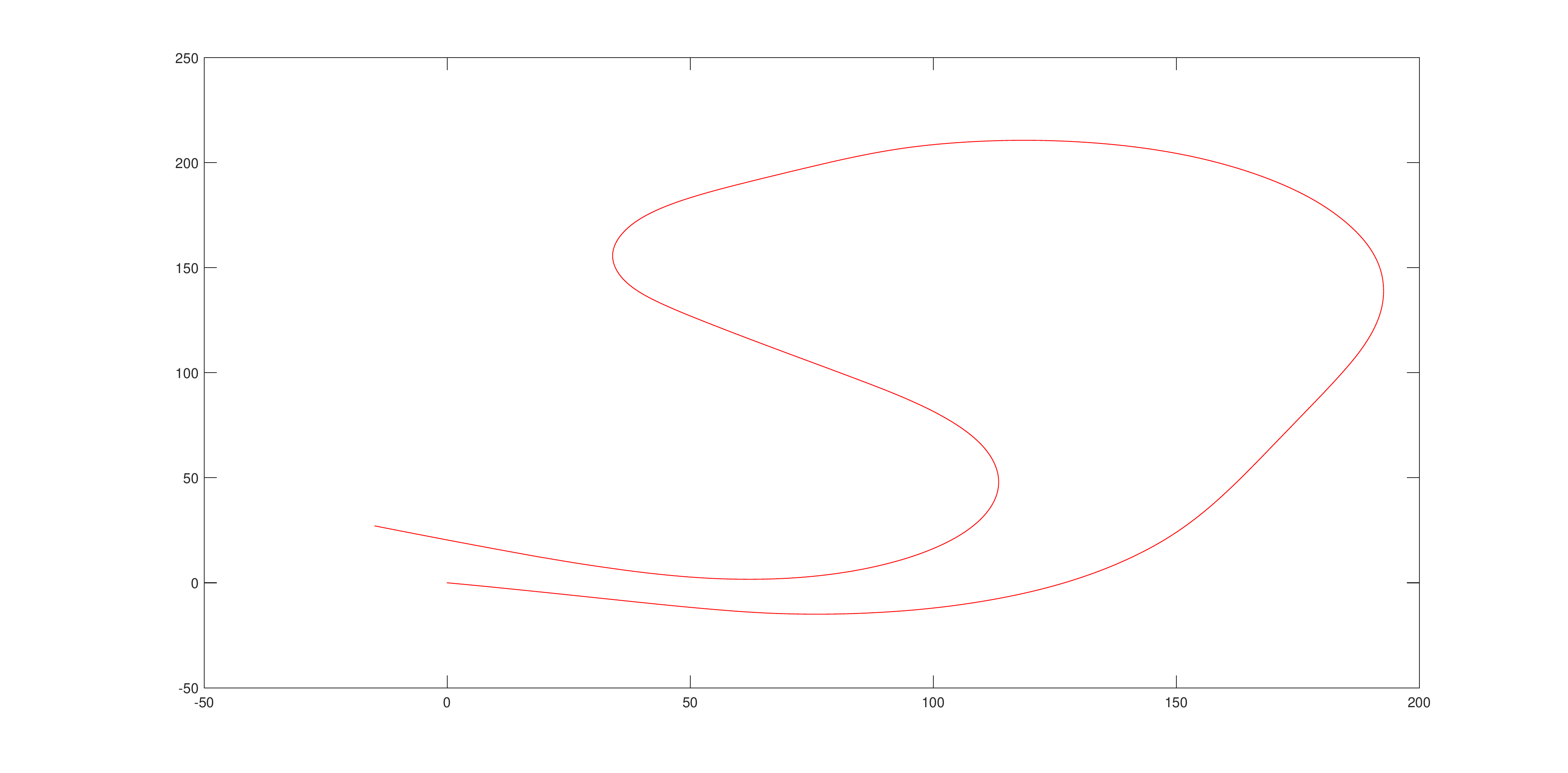}
	\caption{Trajectory of the ego vehicle. The starting point is $(0,0)$.}
\end{figure}
Variations of the relative $x,y$ (in metre) $vx,vy$ (in metre/second) as a function of time (second) in one experience among the seven are displayed in the next 4 figures.
\begin{figure}[H]
	\centering
	\includegraphics[width=8cm,height=5.3cm]{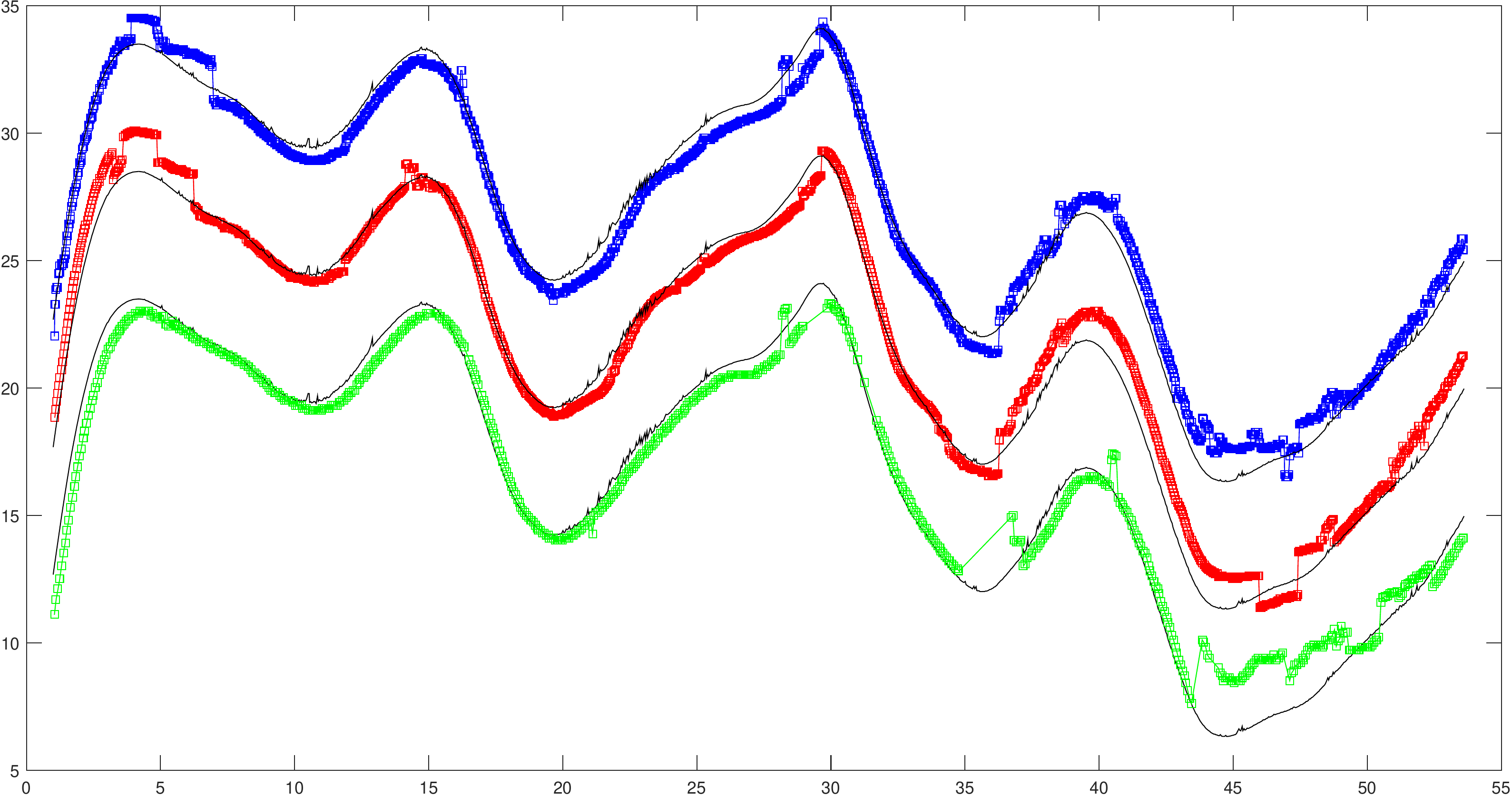}
	\caption{variations of $x$ (in m) with Radar (green)/Lidar (red)/Fusion (blue) in comparison with RTK (black) as a function of time (in s). Offset=5.}
\end{figure}
\begin{figure}[H]
	\centering
	\includegraphics[width=8cm,height=5.3cm]{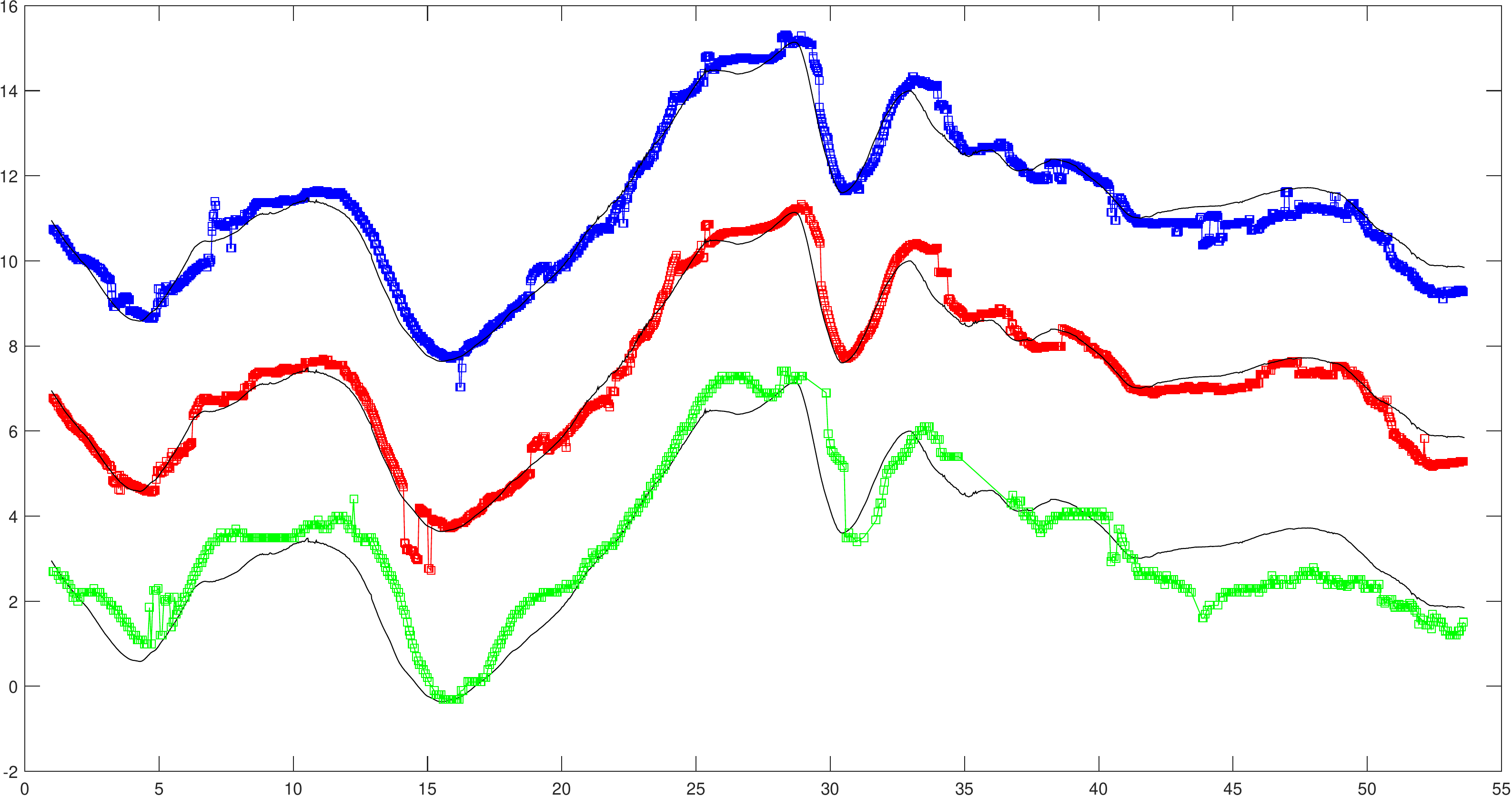}
	\caption{variations of $y$ with Radar/Lidar/Fusion. Offset=4.}
\end{figure}
\begin{figure}[H]
	\centering
	\includegraphics[width=8cm,height=5.3cm]{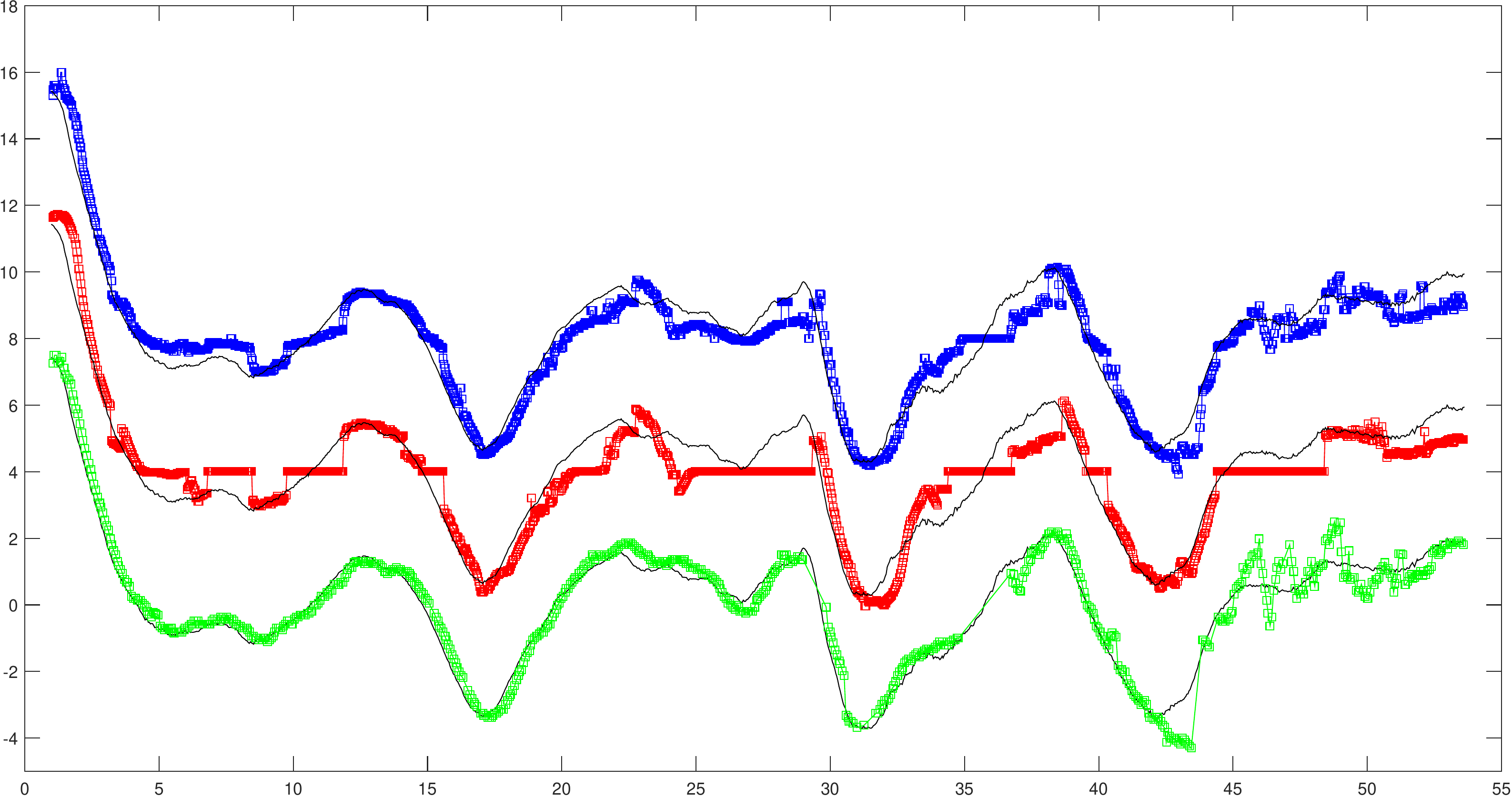}
	\caption{variations of $vx$ with Radar/Lidar/Fusion. Offset=4.}
\end{figure}
\begin{figure}[H]
	\centering
	\includegraphics[width=8cm,height=5.3cm]{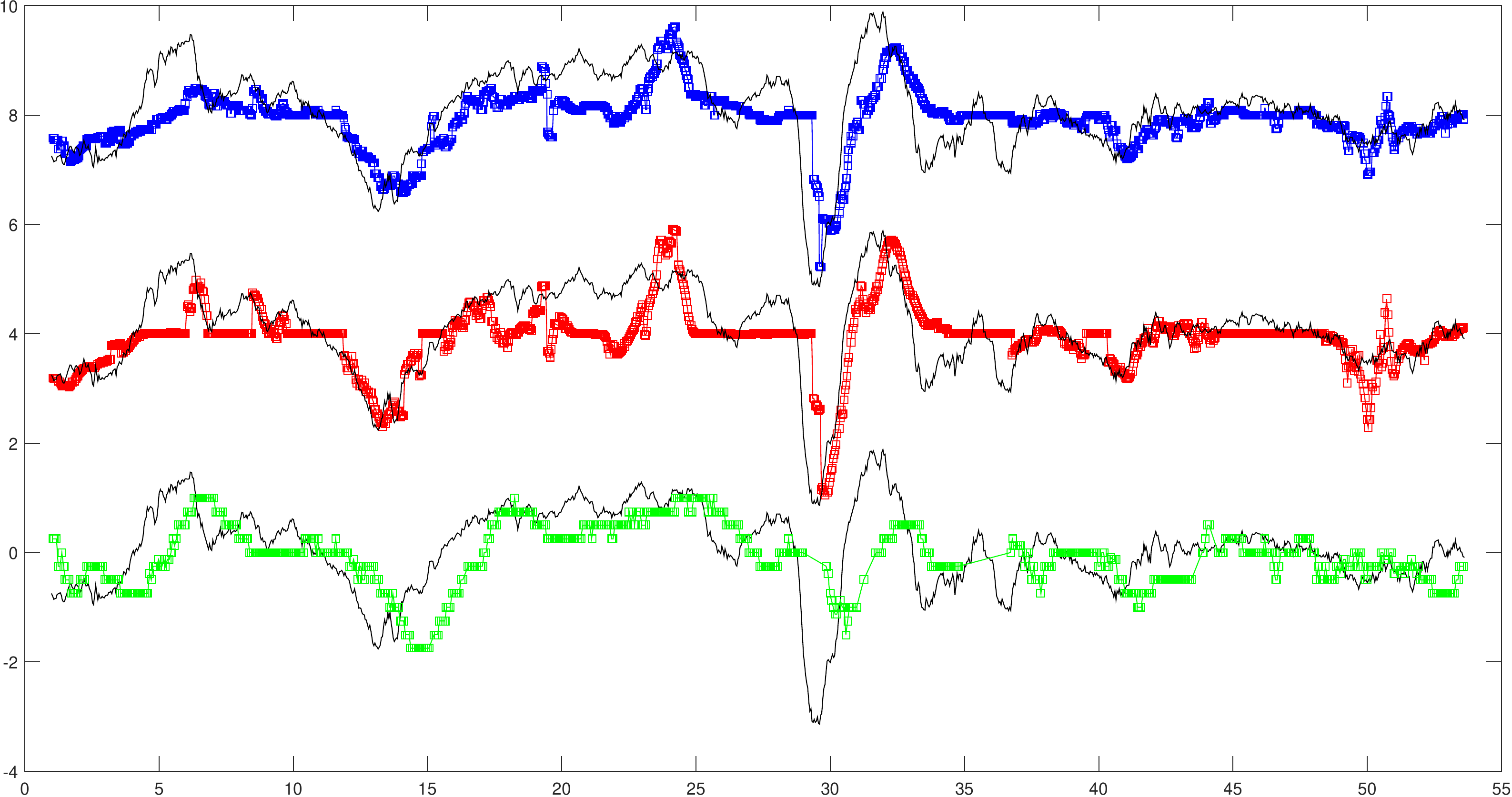}
	\caption{variations of $vy$ with Radar/Lidar/Fusion. Offset=4.}
\end{figure}
Table \ref{tab1} shows the mean square erros (MSE) on $x,y,vx,vy$ by Radar,Lidar,Fusion corresponding to the seven experiences. 
\begin{table}[H]
	\centering
	\begin{tabular}{|l||l|l|l|}
		\hline  & Radar &  Lidar &  Fusion  \\
		\hline  MSE on $x$  & $0.8$ & $0.61$  &$0.65$\\
		\hline  MSE on $y$ & $0.5$ & $0.15$ & $0.19$  \\
		\hline  MSE on $vx$ & $0.16$ & $0.45$ & $0.29$  \\
		\hline  MSE on $vy$ & $0.33$ & $0.38$ & $0.36$  \\
		\hline
	\end{tabular}
	\caption{Table of MSEs of $x,y,vx,vy$ by Radar/Lidar/Fusion.}
	\label{tab1}
\end{table}
\subsection{Comments}
\begin{itemize}
	\item [(a)] The plots of $x$ and $y$ are relatively smooth for Lidar/Radar/Fusion in both scenarios. In contrast, the plots of $vx$ and $vy$ for Lidar present multiple brutal transitions and piecewise constancies. These drawbacks are minimal for Radar and Fusion. 
	\item [(b)] Multiple non detection periods of the target are observed for Radar especially in the second scenario. This can be explained by the fact that the target is not in the field of view of the sensor or also by a sensor malfunction.
	\item [(c)] It is remarkable that Radar was more accurate on $x$ than Lidar in the first scenario. This fact is supported by the experiment presented in \cite{ttt} in which Radar was more accurate than Lidar on the relative distance $=\sqrt{x^2+y^2}$. 	
	\item [(d)] Experiments lead to the following conclusion. First, in terms of accuracy, Fusion provides a good compromise value between Radar and Lidar. Second, Fusion is more robust against unavailability of Lidar and Radar. Unavailability of information has dangerous impact in autonomous driving. This problem is very unlikely to occur for Fusion as the latter combines two sensors. 
\end{itemize}
\subsection{Conclusion and future works}
This paper presented a real-time fusion algorithm between Lidar and Radar which was implemented and successfully integrated into the autonomous car. This algorithm is based on the GNN filter and outputs multiple characteristics of the detected objects such as their relative positions/ velocities and uncertainties. Performance of Fusion in comparison with Radar and Lidar was evaluated through multiple tracking scenarios (on a highway and a bend) using two synchronised vehicles and relying on data coming from ultra-precise RTK sensors as a ground truth. Benefits of Fusion were illustrated through two main central ideas: accuracy regarding the ground truth and robustness against sensors malfunctions. In future works, we plan to conduct experiencies in challenging conditions and use ground truth to compare our approach with the evidence theory approach \cite{Chavez-Garcia2015}. 

\bibliographystyle{IEEEtran}
\bibliography{biblio}

\end{document}